\documentclass[runningheads]{llncs}

\usepackage[T1]{fontenc}
\usepackage[utf8]{inputenc}
\usepackage[english,shorthands=off]{babel}
\usepackage{csquotes}
\usepackage{mathtools,amsmath,dsfont}

\newcommand*{\forexample}{e.g.\ }
\newcommand*{\idest}{i.e.\ }
\newcommand*{\R}{\mathds{R}}
\newcommand*{\ilprule}[1]{\small\ttfamily#1\normalfont\normalsize}

\usepackage{graphicx}
\graphicspath{{./figures/}}

\usepackage[misc]{ifsym} 

\usepackage{booktabs,multirow}
\usepackage[%
table-number-alignment = center,
table-figures-integer = 1,
table-figures-decimal = 3]{siunitx}

\usepackage{algorithm}
\usepackage[noend]{algpseudocode}
\usepackage{comment}
\usepackage{xcolor,hyperref}
\hypersetup{
  pdfauthor={J. Rabold, G. Schwalbe, and U. Schmid},
  pdftitle={Expressive Explanations of DNNs by Combining Concept Analysis with ILP},
}

\begin{document}
\title{Expressive Explanations of DNNs by Combining Concept Analysis with ILP}
%
%
\author{Johannes Rabold \Letter\inst{1}\orcidID{0000-0003-0656-5881}
  \and Gesina Schwalbe\inst{1,2}\orcidID{0000-0003-2690-2478}
  \and Ute Schmid\inst{1}\orcidID{0000-0002-1301-0326}}
\authorrunning{J. Rabold et al.}
%
\institute{
  Cognitive Systems, University of Bamberg, Germany
  \email{\{forename.lastname\}@uni-bamberg.de}
  \and
  Holistic Engineering and Technologies, Artificial Intelligence\\
  Continental AG, Regensburg, Germany\\
  \email{\{forename.lastname\}@continental-corporation.com}
}

\maketitle              
\begin{abstract}

  Explainable AI has emerged to be a key component for black-box machine learning 
  approaches in domains with a high demand for reliability or transparency.
  Examples are medical assistant systems, and applications concerned with the
  General Data Protection Regulation of the European Union, which features
  transparency as a cornerstone.
  Such demands require the ability to audit the rationale behind a classifier's decision.
  While visualizations are the de facto standard of explanations, they
  come short in terms of expressiveness in many ways:
  They cannot distinguish between different attribute manifestations of visual features
  (\forexample eye open vs. closed), and they cannot accurately describe the
  influence of \emph{absence} of, and \emph{relations} between
  features. An alternative would be more expressive symbolic surrogate models.
  However, these require symbolic inputs, which are not readily available in most 
  computer vision tasks.
  In this paper we investigate how to overcome this:
  We use inherent features learned by the network to build a global, expressive,
  verbal explanation of the rationale of a feed-forward convolutional
  deep neural network (DNN).
  The semantics of the features are mined by a concept analysis 
  approach trained on a set of human understandable visual concepts. 
  The explanation is found by an Inductive Logic Programming (ILP) method and 
  presented as first-order rules. We show that our explanation is faithful to the 
  original black-box model%
  \footnote{The code for our experiments is available at\\
    \url{https://github.com/mc-lovin-mlem/concept-embeddings-and-ilp/tree/ki2020}}.

\end{abstract}
%
%
%
\section{Introduction}

Machine learning went through several changes of research perspective
since its beginnings more than fifty years ago. Initially, machine
learning algorithms were inspired by human learning
\cite{Michalski83a}.  Inductive Logic Programming (ILP)
\cite{muggleton1991inductive} and explanation-based generalization
\cite{mitchell1986explanation} were introduced as integrated
approaches which combine reasoning in first-order logic and inductive
learning.

With the rise of statistical approaches to machine learning, focus
shifted from human-like learning to optimizing learning for high
predictive accuracy. Deep learning architectures
\cite{goodfellow2016deep} resulted in data-intensive, black-box
approaches with impressive performances in domains such as object
recognition, machine translation, and game playing. However, since
machine learning more and more is moving from the lab to the real
world, researchers and practitioners alike realize that interpretable,
human-like approaches to machine learning are necessary to allow
developers as well as end-users to evaluate and understand classifier
decisions or possibly also the learned models themselves.

Consequently there is a growing number of approaches to support
explainability of black-box machine learning
\cite{adadi2018peeking}. Explainable AI (XAI) approaches are proposed
to support developers to recognize oversampling and problems with data
quality such as number of available data, class imbalance, expensive
labeling, and sampling biases
\cite{lapuschkin2019unmasking,arya2019one}. For many application
domains, it is a legal as well as an ethical obligation to make
classifier decisions transparent and comprehensible to end-users who
need to make sense of complex information, for instance in medical
diagnosis, automotive safety, or quality control. 

A main focus of research on explanations for image classifications is
on visual explanations, that is,  highlighting of relevant pixels such
as LRP \cite{samek2017explainable} or showing relevant areas in the
image such as LIME \cite{ribeiro2016should}. However, visual
explanations can only show which conjunction of information in an
image is relevant. In many domains, more sophisticated information
needs to be taken into account \cite{Schmid2018}:

\begin{itemize}

\item \textbf{Feature values:}
  highlighting the area of the eye in an image is not helpful to
  understand that it is important for the class decision that the lids
  are tightened (indicating pain) in contrast to eyes which are wide
  open (indicating startle, \cite{weitz2019deep});

\item \textbf{Quantification:}
  highlighting all blowholes on the supporting parts of a rim does not
  make clear that the rim is not a reject because \emph{all} blowholes
  are smaller than 0,5 mm;

\item \textbf{Negation:}
  highlighting the flower in the hand of a person does not transport
  the information that this person is \emph{not} a terrorist because
  he or she does \emph{not} hold a weapon; 

\item \textbf{Relations:}
  highlighting all windows in a building cannot help to discriminate
  between a tower, where windows are \emph{above} each other and a
  bungalow, where windows are \emph{beside} each other
  \cite{rabold2019enriching};

\item \textbf{Recursion:}
  highlighting all stones within a circle of stones cannot transport
  the information that there must be a sequence of an arbitrary number
  of stones with increasing size \cite{rabold2018explaining}.

\end{itemize}  

Such information can only be expressed in an expressive language, for
instance some subset of first-order logic
\cite{muggleton2018ultra}. In previous work, it has been shown how ILP
can be applied to replace the simple linear model agnostic
explanations of LIME
\cite{dai2019bridging,rabold2018explaining,rabold2019enriching,schmid2020mutual}. Alternatively,
it is investigated how knowledge can be incorporated into deep
networks. For example, capsule networks \cite{sabour2017dynamic} are
proposed to model hierarchical relationships and embeddings of
knowledge graphs allow to grasp relationships between entities
\cite{ji2015knowledge}.

In this paper, we investigate how symbolic knowledge can be extracted
from the inner layers of a deep convolutional neural network to
uncover and extract relational information to build an expressive
global explanation for the network. In the following, we first
introduce concept embedding analysis (to extract visual concepts) and
ILP (to build the explanation). In section 3, the proposed approach to
model-inherent generation of symbolic relational explanations 
is presented. We present a variety of experiments on a new ``Picasso''
data set of faces with permuted positions of sub-parts such as eyes,
mouth, and nose. We conclude with an outlook to extend this first,
preliminary investigation in the future. 
\section{Theoretical Background}

\subsection{Concept Embedding Analysis}\label{sec:conceptanalysis}

To understand the process flow of an algorithm, it is of great value
to have access to interpretable intermediate outputs.
The goal of concept embedding analysis is to answer \emph{whether},
\emph{how well}, \emph{how}, and with what
\emph{contribution to the reasoning}
information about semantic concepts is embedded into the latent spaces
(intermediate outputs) of DNNs, and to provide the result in an
explainable way.
Focus currently lies on finding embeddings in either the complete
output of a layer (image-level concepts), or single pixels of an
activation map of a convolutional DNN (concept segmentation).
%
To answer the \emph{whether}, one can try to find a decoder for the
information about the concept of interest, the \emph{concept embedding}.
This means, one is looking for a classifier on the latent space that
can predict the presence of the concept.
The performance of the classifier provides a measure of \emph{how well}
the concept is embedded.
%
For an explainable answer of \emph{how} a concept is embedded, the
decoder should be easily interpretable.
One constraint to this is introduced by the rich vector space
structure of the space of semantic concepts respectively word vector spaces
\cite{mikolov_linguistic_2013}:
The decoder map from latent to semantic space should preserve at least
a similarity measure.
For example, the encodings of \enquote{cat} and \enquote{dog} should
be quite similar, whereas that of a \enquote{car} should be relatively
distant from the two.
The methods in literature can essentially be grouped by their choice
of distance measure $\langle -,-\rangle$ used on the latent vector space.
A concept embedding classifier $E_c$ predicting the presence of concept $c$
in the latent space $L$ then is of the form
$E_c(v)=\langle v_c,v\rangle > t_c$
for $v\in L$,
where $v_c\in L$ is the concept vector of the embedding,
and $t_c\in \R$.

Automated concept explanations \cite{ghorbani_towards_2019} uses
$L_2$ distance as similarity measure. They discover concepts in an
unsupervised fashion by k-means clustering of the latent space
representations of input samples. The concept vectors of the
discovered concepts are the cluster centers.
%
In TCAV~\cite{kim_interpretability_2018} it is claimed that the
mapping from semantic to latent space should be linear for best
interpretability. To achieve this, they suggest to use linear
classifiers as concept embeddings. This means they try to find a
separation hyperplane between the latent space representations of
positive and negative samples of the concept.
A normal vector of the hyperplane then is their concept vector,
and the distance to the hyperplane
is used as distance measure.
As method to obtain the embedding they use support vector machines (SVMs).
TCAV further investigated the contribution of concepts to given output
classes by sensitivity analysis.
A very similar approach to TCAV, only instead relying on logistic
regression, is followed by Net2Vec~\cite{fong_net2vec_2018}.
As a regularization, they add a filter-specific cut-off before the
concept embedding analysis to remove noisy small activations.
The advantage of Net2Vec over the SVMs in TCAV is that they can
more easily be used in a convolutional setting: They used a
1$\times$1-convolution to do a prediction of the concept for each
activation map pixel, providing a segmentation of the concept.
This was extended by \cite{schwalbe_concept_2020}, who suggested to
allow larger convolution windows
to ensure that the receptive field of the window can cover
the complete concept. This avoids a focus on local patterns.
%
A measure that can be applied to concept vectors of the same layer
regardless of the analysis method, is that of
\emph{completeness} suggested in \cite{yeh_completeness-aware_2020}.
They try to measure, how much of the information relevant to the final
output of the DNN is covered by a chosen set of concepts vectors.
They also suggested a metric to compare the attribution of each
concept to the completeness score of a set of concepts.

\subsection{Inductive Logic Programming}\label{sec:ilp}

Inductive Logic Programming (ILP)~\cite{muggleton1991inductive} is a
machine learning technique that builds a logic theory over positive
and negative examples ($E^+$, $E^-$). The examples consist of symbolic
background knowledge (BK) in the form of first-order logic predicates,
\forexample \ilprule{contains(Example, Part), isa(Part, nose)}. Here the
upper case symbols are variables and the lower case symbol is a
constant. The given BK describes that example \ilprule{Example}
contains a part \ilprule{Part} which is a nose. Based on the examples,
a logic theory can be learned. The hypothesis language of this theory
consists of logic Horn clauses that contain predicates from the BK. We
write the Horn clauses as implication rules,
\forexample
\ilprule{
  \begin{align*}
    \text{face(Example) :- }& \text{contains(Example, Part), isa(Part, nose)}\;.
  \end{align*}
}
For this work we obey the syntactic rules of the Prolog
programming language. The \ilprule{:-} denotes the logic implication
($\leftarrow$). We call the part before the implication the
\emph{head} of a rule and the part after it the \emph{body} or
\emph{preconditions} of a rule. 

We use the framework Aleph~\cite{srinivasan2001aleph}
for this work since it is a flexible and adaptive general purpose ILP
toolbox. Aleph's built in algorithm attempts to induce a logic theory
from the given BK to cover as many positive examples $E^+$ as possible
while avoiding covering the negative examples $E^-$. The general
algorithm of Aleph can be summarized as
follows~\cite{srinivasan2001aleph}:

\begin{enumerate}
\item As long as positive examples exist, select one. Otherwise halt.
\item Construct the most-specific clause that entails the selected example and is within the language constraints.
\item Find a more general clause which is a subset of the current literals in the clause.
\item Remove examples covered by the current clause.
\item Repeat from step 1.
\end{enumerate}

\section{Explaining a DNN with Concept Localization and ILP}

When building explanations for a DNN via approximate rule sets,
the underlying logic and predicates of the rules should reflect the
capabilities of the model.
For example, spatial relations like \ilprule{top\_of} or
\ilprule{right\_of} should be covered, as \forexample dense
layers of a DNN are capable of encoding these.
Spatial relations cannot be represented by current visualization
methods for explainable AI, which only feature predicates of the form
\ilprule{contains(Example, Part)} and \ilprule{at\_position(Part, xy)}.
Rule-based methods like ILP are able to incorporate richer
predicates into the output. However, their input must be symbolic background
knowledge about the training and inference samples which
is formulated using these predicates \emph{explicitly}.
For computer vision tasks with pixel-level input, this
encoding of the background knowledge about samples is not available.
To remedy this, we propose to use existing concept mining techniques
for extraction of the required background knowledge:
\begin{enumerate}
\item Associate pre-defined visual semantic concepts with
  intermediate output of the DNN.
  Concepts can be local, like parts and textures, or image-level.
\item Automatically infer the background knowledge about a sample
  \ilprule{Ex} given the additional concept output, which defines
  predicates
  \ilprule{isa(C, concept)}, and
  \ilprule{isa(Ex, C)} (image-level) or
  \ilprule{contains(Ex, C)} with \ilprule{at\_position(C, xy)}.
  From this, spatial relations and negations can be extracted.
\item Given background knowledge for a set of training samples,
  apply an inductive logic programming approach to learn an
  expressive set of rules for the DNN.
\end{enumerate}

The approach presented in this paper differs from the previous work
outlined in \cite{rabold2019enriching} by the following main aspects:

\begin{itemize}
\item We will find a global verbal explanation for a black-box
  decision in contrast to a local explanation.
\item We directly make use of information stored in the building
  blocks of the DNN instead of relying on the linear surrogate
  model generated by LIME.
\end{itemize}

\subsection{Enrich DNN Output via Concept Embedding Analysis}
We directly built upon the concept detection approach
from \cite{fong_net2vec_2018,schwalbe_concept_2020}, suggesting some
further improvements.
Net2Vec bilinearly upscaled the predicted masks before
applying the sigmoid for logistic regression.
This overrates the contribution to the loss by pixels at
the edges from positive to negative predicted pixels. We instead
apply upscaling after applying the sigmoid.
%
Instead of the suggested IoU penalty from
\cite{schwalbe_concept_2020}, we propose a more stable
Dice loss to fit the overlap objective, supported by a small summand
of the balanced binary cross-entropy (bBCE) suggested in Net2Vec to
ensure pixel-wise accuracy.

One major disadvantage of the linear model approaches over the
clustering ones is their instability, \idest several runs for the same
concept yield different concept vectors.
Reasons may be dependence on the outliers of the concept cluster
(SVM); non-unique solutions due to a margin between the clusters;
and inherent variance of the used optimization methods.
To decrease dependence on the training set selection and ordering, and
the initialization values, we for now simply use ensembling.
For this we define a hyperplane $H$ as the zero set of the distance function
$d_H(v) = (v - b_H\cdot v_H) \circ v_H$
for the normal vector $v_H$ and the support vector $b_Hv_H$, $b_H\in\R$.
Then, the zero set of the mean
$\frac{1}{N} \sum_{i=1}^{N} d_{H_i}$ of the distance functions of
hyperplanes $H_i$ again defines a hyperplane with
\begin{align*}
  v_H &= \textstyle\frac{1}{N} \sum_{i=1}^{N} v_{H_i} &\text{and}&
  &b_H &= \textstyle\frac{1}{\|v_H\|^2} \frac{1}{N} \sum_{i=1}^{N} (b_{H_i}\|v_{H_i}\|^2) \;.
\end{align*}
Note, that hyperplanes with longer normal vectors (\idest higher
confidence values) are overrated in this calculation. To remedy this,
concept vectors are normalized before ensembling, using the property
$(w - b\cdot v) \circ v
= \|v\| \cdot (w - (b\|v\|)\frac{v}{\|v\|})\cdot \frac{v}{\|v\|}$
of the distance function for scalar $b$ and vectors $v, w$.

\subsection{Automatic Generation of Symbolic Background Knowledge}
\label{sec:bkextraction}

The output of the concept analysis step (binary masks indicating the
spatial location of semantic concepts) can be used to build a symbolic
global explanation for the behavior of the original black-box
model. We obtain the explanation by finding a first-order logic theory
with the ILP approach Aleph (see Section~\ref{sec:ilp}).
Since Aleph needs a set of positive and negative examples
($E^+$, $E^-$), the first step is to obtain these examples along with their
corresponding symbolic background knowledge (BK). In order to obtain a
good approximation of the behavior of the model, we sample $N^+$
binary masks from positively predicted images and $N^-$ binary masks
from negatively predicted images that lie close to the decision
boundary of the original black-box model using the concept analysis
model described above. Let $M^+$, $M^-$ be the set of positive and
negative binary masks. Let $m^+ \in M^+$, $m^- \in M^-$ be single
masks. Each mask (\forexample $m^+$) consists of multiple mask layers
(\forexample $l_c \in m^+$, $c \in C$) for the different human
understandable concepts from the pool of concepts $C$. These mask
layers are sparse matrices upsampled to the same size as the original
images they are masking. The matrices have the value 1 at all the
positions where the concept analysis model detected the respective
concept and 0 at all other positions.

The symbolic explanation of the original model should consist of logic
rules that establish the prototypical constellation of visual parts of
an image that resembles the positive class as seen by the DNN. We
therefore need not only the information about occurrence of certain
visual parts in the sampled examples but also the different relations
that hold between the parts. In the next sections we adhere to the
following general workflow:

\begin{enumerate}
\item Find positions of visual parts in the examples and name them.
\item Find relations between parts.
\item Build BK with the information from step 1 and 2.
\item Induce a logic theory with Aleph.
\end{enumerate}

\subsubsection{Find Visual Parts}

One mask layer $l_c$ contains possibly multiple contiguous clusters of
concept propositions. Therefore, as a denoising step, we only take the
cluster with the largest area into account. As a proposition for the
position of the concept $c$ in the picture, in this cluster we take
the mean point of the area $A_{max}$ of 1's in $l_c$. We therefore
find the position $p_c = (x, y)$ with $x = (\min_x(A_{max}) +
\max_x(A_{max})) / 2$ and $y = (\min_y(A_{max}) + \max_y(A_{max})) / 2$.
This procedure can be followed for all masks $l_c$ that are
contained in all $m^+ \in M^+$ and $m^- \in M^-$.


\subsubsection{Find Relations Between Parts}

\begin{figure}[t]
  \centering
  \includegraphics[height=3cm]{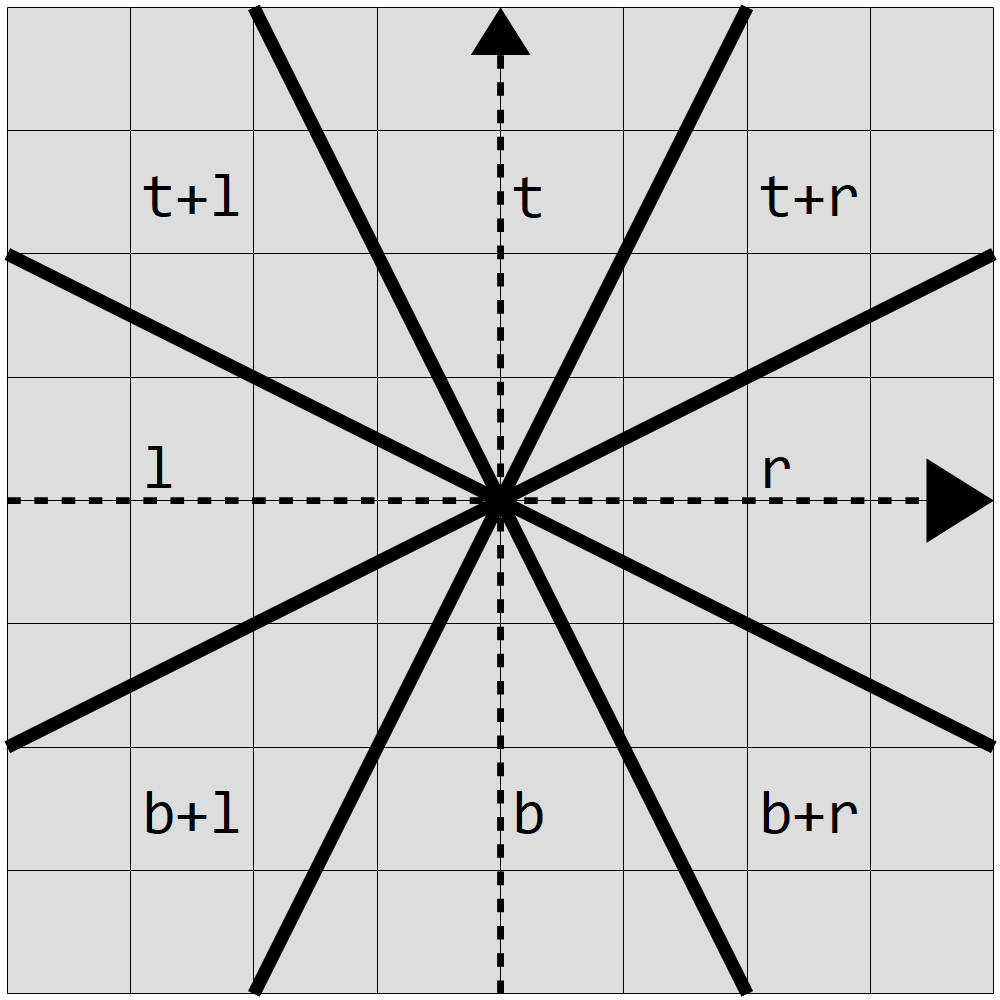}
  \caption{Potential positions of an object to be only \textbf{t}op
    of, \textbf{r}ight of, \textbf{b}ottom of, or \textbf{l}eft of a
    reference object located in the origin. Also the four
    overlapping regions are indicated.}
  \label{fig:relations}
\end{figure}

By taking relationships between the parts into account, we strive for
more expressive explanations. For this work we limit ourselves to
spatial relationships that hold between the parts that were found in
the previous steps. We assume that pairs of two parts can be in the
following four relationships to each other: \ilprule{left\_of},
\ilprule{right\_of}, \ilprule{top\_of}, \ilprule{bottom\_of}. We
declare part \ilprule{A} to be \ilprule{top\_of} part \ilprule{B} if
the vertical component $y_A$ of the position $p_A$ is above $y_B$ and
the value for the horizontal offset $\Delta x = x_A - x_B$ does not
diverge from the value for the vertical offset $\Delta y = y_A - y_B$
by more than double. The other spatial relations can be formalized in
an analogous manner. Thus, the relations that can hold between two
parts \ilprule{A} and \ilprule{B} can be visualized as in
Fig.~\ref{fig:relations}.

\subsubsection{Inferring Global Symbolic Explanations}

After the inference of visual parts and the relationships that hold
between them, we can build the BK needed for Aleph. Part affiliation
to an example can be declared by the \ilprule{contains} predicate. We
give all parts a unique name over all examples. Suppose part
\ilprule{A} is part of a particular example \ilprule{E} and describes
the human understandable concept $\text{\ilprule{c}} \in C$. Then the
example affiliation can be stated by the predicates
\ilprule{contains(E, A)}, \ilprule{isa(A, c)}.
Likewise for the relations we can use 2-ary predicates that state the
constellation that holds between the parts. When part \ilprule{A} is
left of part \ilprule{B} we incorporate the predicate
\ilprule{left\_of(A, B)} in the BK and likewise for the other
relations. When the BK for the positive and negative examples $E^+$
and $E^-$ is found, we can use Aleph's induction mechanism to find the
set of rules that best fit the examples. The complete algorithm for
the process is stated in Algorithm~\ref{alg:ex_gen}.

\begin{algorithm}[t]
  \caption{\label{alg:ex_gen} Verbal Explanation Generation for DNNs}
  \begin{algorithmic}[1]
    \State \textbf{Require:} Positive and negative binary masks $M^+$, $M^-$
    \State \textbf{Require:} Pool of human understandable concepts $C$
    
    \State $E^+ \gets \{\}$
    \State $E^- \gets \{\}$
    \State \textbf{for each} $\odot \in \{+, -\}$ \textbf{do}
    \State ~~~\textbf{for each} $m^\odot \in M^\odot$ \textbf{do}
    \State ~~~~~~$P \gets \{\}$
    \State ~~~~~~\textbf{for each} $l_c \in m^\odot$ where $c \in C$ \textbf{do}
    \State ~~~~~~~~~$p_c \gets calculatePartPosition(l_c)$
    \State ~~~~~~~~~$P \gets P \cup \{\langle c, p_c \rangle\}$
    \State ~~~~~~$R \gets calculateRelations(P)$
    \State ~~~~~~$E^\odot \gets E^\odot \cup {\langle P, R \rangle}$
    \State $T \leftarrow$ Aleph$(E^+, E^-)$
    \State \textbf{return} $T$
  \end{algorithmic}
\end{algorithm}

\section{Experiments and Results}

\begin{figure}
  \begin{center}
    \includegraphics[width=.15\textwidth]{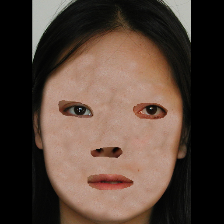}\quad
    \includegraphics[width=.15\textwidth]{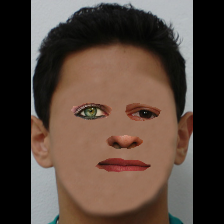}\qquad\qquad
    \includegraphics[width=.15\textwidth]{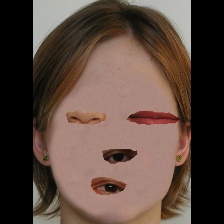}\quad
    \includegraphics[width=.15\textwidth]{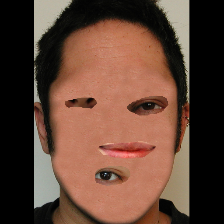}
  \end{center}
  \caption{Examples from the Picasso Dataset
    (\textit{left:} positive class, \textit{right:} negative).}
  \label{fig:picasso}
\end{figure}

We conducted a variety of experiments to audit our previously
described approach. As a running example we used a DNN which we
trained on images from a generated dataset we dubbed \emph{Picasso
  Dataset}. The foundation are images of human faces we took from the
FASSEG dataset~\cite{khan2015multi,khan2017head}.
The Picasso Dataset contains collage images of faces with the facial
features (eyes, mouth, nose) either in the correct constellation
(positive class) or in a mixed-up constellation (negative class). See
Fig.~\ref{fig:picasso} for examples.
No distinction is made between originally left and right eyes.

In order to not establish a divergence in the image space of the two
classes, the positive and negative classes contain facial features
that were cut out of a set of images from the original FASSEG
dataset. As a canvas to include the features, we took a set of
original faces and got rid of the facial features by giving the
complete facial area a similar skin-like texture. Then we included the
cut out facial features onto the original positions of the original
features in the faces.

The face images in Fig.~\ref{fig:picasso} show that
the resulting dataset is rather constructed. This however will suffice for
a proof of concept to show that our approach in fact exploits object parts
and their relations. In the future we plan on moving towards more natural
datasets.

\subsection{Analyzed DNNs}
We evaluated our method on three different architectures from the pytorch
model-zoo\footnote{\url{https://pytorch.org/docs/stable/torchvision/models.html}}:
AlexNet~\cite{krizhevsky_one_2014},
VGG16~\cite{simonyan_very_2015}, and
ResNeXt-50~\cite{xie_aggregated_2017}.
The convolutional parts of the networks were initialized with weights
pre-trained on the ImageNet dataset.
For fine-tuning the DNNs for the Picasso Dataset task,
the output dimension was reduced to one and the in- and output
dimension of the second to last hidden dense layer was reduced to 512
for AlexNet and VGG16.
Then the dense layers and the last two convolutional layers
(AlexNet, VGG16) respectively bottleneck blocks (ResNeXt) were fine-tuned.
The fine-tuning was conducted in one epoch on a training set of 18,002
generated, $224\times224$-sized picasso samples with equal
distribution of positive and negative class.
All models achieved accuracy greater than 99\%
on a test set of 999 positive and 999 negative samples.


\subsection{Training the Concept Models}
\begin{figure}%
  \centering%
  \hspace*{-0.08\textwidth}\includegraphics[width=1.16\textwidth]{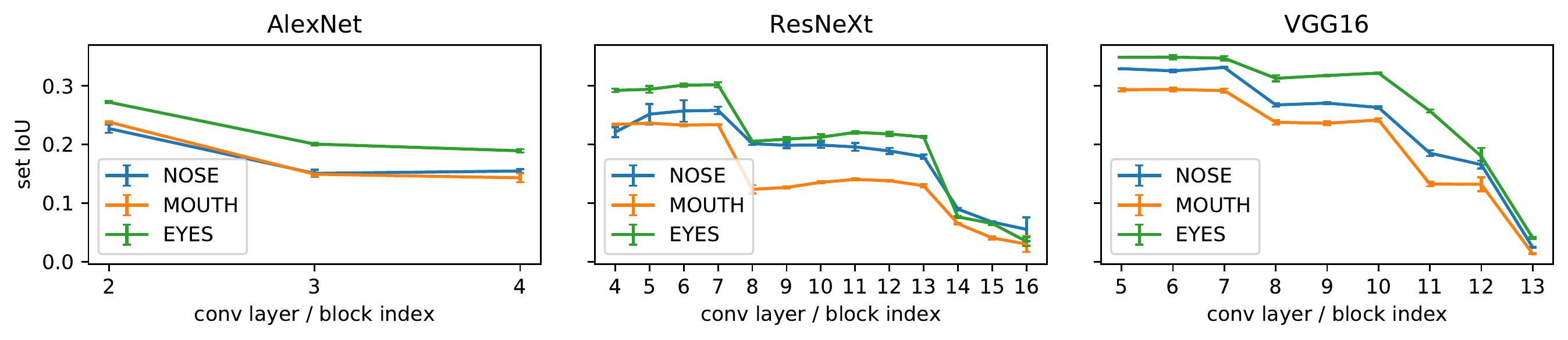}
  \caption{The layer-wise mean set IoU results of the concept analysis runs.}
  \label{fig:layerwiseresults}
\end{figure}

In our example we determined the best ensembled detection concept vectors for
the concepts \textsc{eyes}, \textsc{mouth} and \textsc{nose} amongst the
considered layers.
%
We excluded layers with low receptive field, as they are assumed to
hold only very local features
(for the layers used see Fig.~\ref{fig:layerwiseresults}).
Convolutional output was only considered after the activation.
%
For each concept, 452 training/validation, and 48 test picasso samples
with segmentation masks were used.
The training objective was: Predict at each activation map pixel
whether the kernel window centered there lies \enquote{over}
an instance of the concept. Over meant that the fuzzy intersection of
the concept segmentation and the kernel window area exceeds a threshold
(\emph{intersection encoding}).
This fuzzy definition of a box center
tackles the problem of sub-optimal intersections in later layers
due to low resolution.
Too high values may lead to elimination of an instance,
and thresholds were chosen to avoid such issues with
values 0.5/0.8/0.7 for nose/mouth/eye.
We implemented the encoding via a convolution.
As evaluation metric we use set IoU (sIoU) between the detection
masks and the intersection encoded masks as in Net2Vec.
%
On each dataset and each layer, 15 concept models were trained in
three 5-fold-cross-validation runs with the following settings:
Adam optimization with mean best learning rate of 0.001,
a weighting of 5:1 of Dice to bBCE loss,
batch size of 8, and
two epochs (all layers showed quick convergence).
                  %

\begin{figure}%
  \centering%
  \includegraphics[width=0.6\textwidth]{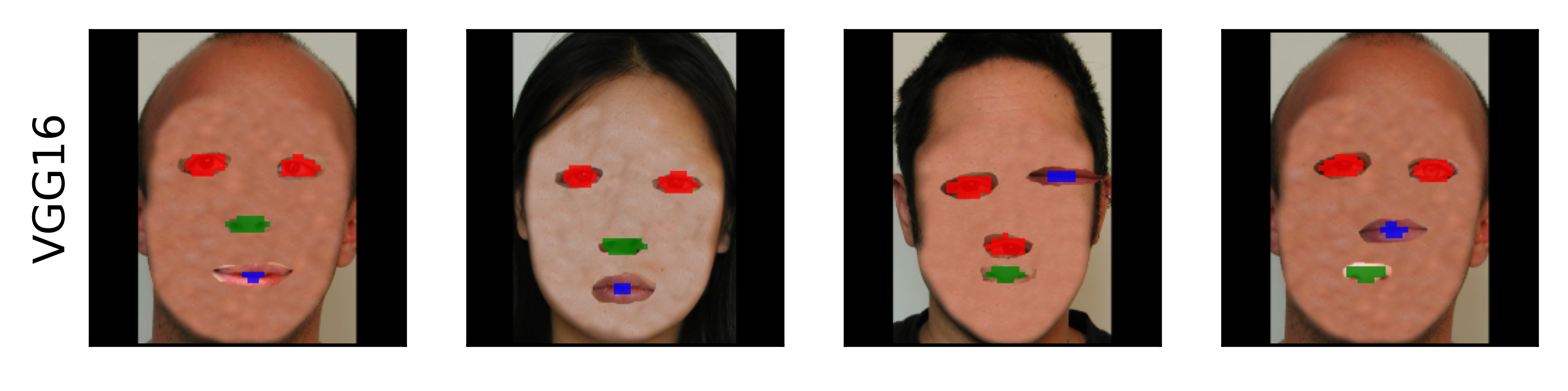}\\
  \includegraphics[width=0.6\textwidth]{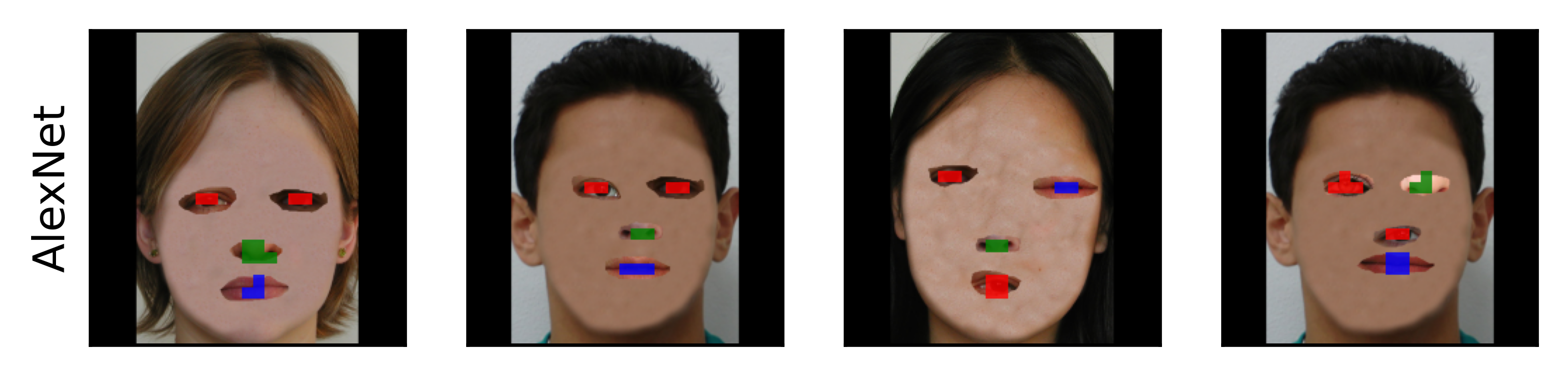}%
  \caption{
    Ensemble embedding outputs of
    \textsc{nose} (green), \textsc{mouth} (blue), \textsc{eyes} (red).}
  \label{fig:conceptsamples}
\end{figure}

\paragraph{Results}
Our normalized ensembling approach proved valuable as it yielded
mean or slightly better performance compared to the single runs.
For the considered models, meaningful embeddings of all concepts could
be found (see Tab.~\ref{tab:bestembeddings}): The layers all reached
sIoU values greater than 0.22
despite of the still seemingly high influence of sub-optimal
resolutions of the activation maps.
Fig.~\ref{fig:conceptsamples} shows some exemplary outputs.
The concepts were best embedded in earlier layers,
while different concepts did not necessarily share the same layer.
\begin{table}
  \centering
  \caption{Results for ensemble embeddings with
    set IoU (sIoU), mean cosine distance to the runs (Cos.d.),
    and index of conv layer or block (L) (cf. Fig.~\ref{fig:layerwiseresults}).
    \label{tab:bestembeddings}}%
  \footnotesize%
  \newcommand*{\normentry}[1]{\multicolumn{1}{c}{#1}}%
  \newcommand{\model}[1]{\multirow{4}{*}{
      \begin{tabular}{@{}c@{}}
        \rotatebox[origin=c]{90}{\strut#1}
      \end{tabular}}}%
  \newenvironment{statstable}[1]{%
    \begin{tabular}{@{}l >{\scshape}l c@{~~} S[table-auto-round] S[table-auto-round]@{}}%
      \toprule%
      \model{#1}
      & 
      & \normentry{L~~}
      & \normentry{sIoU}
      & \normentry{Cos.d.}\\%
      \cmidrule[\heavyrulewidth]{2-5}
    }{\bottomrule\end{tabular}}%
  \mbox{%
    \begin{statstable}{AlexNet}
      & nose  & 2 & 0.227925 & 0.04030534426371257 \\ 
      & mouth & 2 & 0.239023 & 0.04018515348434448 \\ 
      & eyes  & 2 & 0.271918 & 0.0581478198369344  \\ 
    \end{statstable}\hspace*{1em}%
    \begin{statstable}{VGG16}
      & nose  & 7 & 0.332171 & 0.10435034632682805 \\ 
      & mouth & 6 & 0.295552 & 0.15369088053703306 \\ 
      & eyes  & 6 & 0.349707 & 0.1969874024391174  \\ 
    \end{statstable}\hspace*{1em}%
    \begin{statstable}{ResNeXt}
      & nose  & 6 & 0.263708 & 0.017367599999999928 \\ 
      & mouth & 5 & 0.236788 & 0.020498533333333402 \\ 
      & eyes  & 7 & 0.302339 & 0.019860466666666854 \\ 
    \end{statstable}%
  }
\end{table}

\subsection{Example Selection for ILP Training}
The goal of the ILP model is to approximate the behavior of the main
DNN, \idest its decision boundary.
For this, few but meaningful training samples and their DNN output
are needed: class-prototypes as well as ones that tightly frame the
DNN decision boundary.
From the 1,998 samples in the picasso test set, in total 100 samples
were chosen from the DNN test set to train the ILP model.
The DNN confidence score here was used to estimate
the proximity of a data point to the decision boundary.
For each class, we selected the 50 samples predicted to be in this
class and with confidence closest to the class boundary of 0.5.
In our setup this provided a wide range of confidence values
(including 0 and 1).

\subsection{Finding the Symbolic Explanation}
In order to find the background knowledge needed for Aleph to generate
the explanation, we need to extract the information about the facial
features and their constellations from the masks of the samples drawn in the
previous step. Abiding the procedure described in
Sec.~\ref{sec:bkextraction}, we first find contiguous clusters in the
mask layers to then infer the positional information for them. This is
straight-forward for the nose and the mouth but imposes a problem for
the eyes, since we do not want to have a single position proposal for
them in the eye that produces the biggest cluster in the mask
layer. Thus, we allow for the top two biggest clusters to infer a
position. Although we give them unique constants in the BK, we both
give them the type \ilprule{eye} $\in C$.

The next step consists of the extraction of the spatial features
between the found parts. Since the relation pair
\ilprule{left\_of}\,/ \ilprule{right\_of} as well as
\ilprule{top\_of}\,/ \ilprule{bottom\_of} can be seen as the inverses of
the respective other relation, we omit the relations
\ilprule{right\_of} and \ilprule{bottom\_of} in the BK. This is
possible, because the \emph{Closed World Assumption} holds (Everything
that is not stated explicitly is false).

Once the BK is found for all examples, we can let Aleph induce a
theory of logic rules. Consider the induced theory for the trained
VGG16 network:
\ilprule{
  \begin{align*}
    \text{face(F) :- } &\text{contains(F, A), isa(A, nose), contains(F, B), isa(B, mouth),}\\
                       &\text{top\_of(A, B), contains(F, C), top\_of(C, A).}
  \end{align*}
}
The rule explicitly names the required facial concepts \ilprule{nose}
and \ilprule{mouth} and the fact that the nose has to be above the
mouth in order for an image to be a face. Further there is another
unnamed component \ilprule{C} required which has to be placed above
the nose. By construction this has to be one of the eyes. The rule
makes sense intuitively as it describes a subset of correct
constellations of the features of a human face.

To further test the fidelity of the generated explanations to the
original black-box network, we calculated several performance metrics
for a test set of 1998 test images (999 positive and 999 negative
examples). We handled the learned explanation rules as binary
classification model for the test images in BK representation. If an
image representation is covered by the explanation rules, it is
predicted to be positive, otherwise negative. We now can handle the
binary output of the black-box model as ground truth to our
explanation predictions. The performance metrics together with the
induced explanation rules for several DNN architectures are listed in
Tab.~\ref{tab:ilp_results}. It can be seen that the explanations stay
true to the original black-box model.

\begin{table}[t]
  \newcommand*{\rulepart}[1]{\ilprule{\scriptsize #1}}
  \caption{Learned rules for different architectures and their
    fidelity scores (accuracy and F1 score wrt.\ to the original model
    predictions).
    Learned rules are of common form
    \rulepart{face(F) :- contains(F, A), isa(A, nose), contains(F, B), isa(B, mouth), distinctPart}
  }
  \label{tab:ilp_results}
  \begin{tabular*}{\textwidth}{lcc@{\extracolsep{\fill}}l}
    \toprule
    \textbf{Arch.}   & \textbf{Accuracy} & \textbf{F1} & \textbf{Distinct rule part}\\
    \midrule[\heavyrulewidth]
    VGG16   & 99.60\%  & 99.60\%  & \rulepart{top\_of(A, B), contains(F, C), top\_of(C, A)} \\
    AlexNet & 99.05\%  & 99.04\%  & \rulepart{contains(F, C), left\_of(C, A), top\_of(C, B), top\_of(C, A)} \\
    ResNext & 99.75\%  & 99.75\%  & \rulepart{top\_of(A, B), contains(F, C), top\_of(C, A)}\\
    \bottomrule
  \end{tabular*}
\end{table}


\section{Conclusion and Future Work}

Within the described simple experiment we showed that expressive,
verbal surrogate models with high fidelity can be found for DNNs using
the developed methodology. We suggest that the approach is promising
and worth future research and optimization.

%
The proposed concept detection approach requires a concept to have
little variance in its size. It should easily extend to
a concept with several size categories (\forexample close by and far away
faces) by merging the result for each category.
%
A next step for the background knowledge extraction would be to extend
it to an arbitrary number of concept occurences per image, where
currently the algorithm assumes a fixed amount (exactly one \ilprule{mouth}, one
\ilprule{nose}, two \ilprule{eyes}). This could \forexample be achieved by allowing
a maximum number per sliding window rather than an exact amount per image.
%
In cases, where the predicates cannot be pre-defined, one can learn
the relations as functions on the DNN output from examples as
demonstrated in \cite{donadello_logic_2017}.

We further did not consider completeness
(cf. Sec.~\ref{sec:conceptanalysis}) of the chosen concepts:
They may not be well aligned with the decision relevant features used
by the DNN, infringing fidelity of the surrogate model.
We suggest two ways to remedy this:
One could rely on (possibly less interpretable) concepts found via
concept mining \cite{ghorbani_towards_2019}.
Or, since ILP is good at rejecting irrelevant information, one can
start with a much larger set of pre-defined, domain related concepts.
We further
assume that best fidelity can only be achieved with the \emph{minimal}
complete sub-set of most decision-relevant concepts, which fosters
uniqueness of the solution.
For a decision relevance measure see \forexample
\cite{ghorbani_towards_2019}.

It may be noted that the presented concept analysis approach is not
tied to image classification: 
As long as the ground truth for concepts in the form of masks or
classification values is available, the method can be applied to any
DNN latent space
(imagine \forexample audio, text, or video classification).
However, spatial or temporal positions and relations are currently
inferred using the receptive field information of convolutional DNNs.
This restriction may again be resolved by learning of relations.

Lastly, in order to examine the understandability of the induced
explanation in a real world scenario, we need to let explanations be
evaluated in a human user study. For this matter, subjective
evaluation measures have to be specifically designed for verbal
explanations.


\bibliographystyle{splncs04}
\bibliography{literature}

\end{document}